\pdfoutput=1

\documentclass[11pt]{article}

\usepackage[final]{acl}

\usepackage{inconsolata}
\usepackage[T1]{fontenc}
\usepackage{times}
\usepackage{latexsym}
\usepackage{hyperref}
\usepackage{inconsolata}
\usepackage{url}
\usepackage{amsmath}
\usepackage{amsthm}
\usepackage{amsfonts}
\theoremstyle{definition}

\usepackage{graphicx}
\usepackage{subcaption}
\usepackage{booktabs}
\usepackage{multirow}
\usepackage{makecell}
\usepackage{wrapfig}
\usepackage{enumitem}
\usepackage{comment}
\usepackage{blindtext}
\usepackage{xcolor}
\usepackage{svg}
\usepackage{amsfonts}
\usepackage{xspace}
\newcommand{\tulu}{\textsc{T\"ulu}\xspace}

\newcommand\modelname{{\usefont{T1}{Discognate}{m}{n}{DTM}}\xspace}

\usepackage[T1]{fontenc}

\usepackage[utf8]{inputenc}

\usepackage{microtype}

\usepackage{inconsolata}

\title{Disperse-Then-Merge: \\Pushing the Limits of Instruction Tuning via Alignment Tax Reduction}

\author{Tingchen Fu\textsuperscript{\rm 1,2}\footnotemark[2], Deng Cai\textsuperscript{\rm 2}\footnotemark[1]\ , Lemao Liu\textsuperscript{\rm 3}, Shuming Shi\textsuperscript{\rm 2} Rui Yan\textsuperscript{1}\footnotemark[1] \\
\textsuperscript{1}Gaoling School of Artificial Intelligence, Renmin University of China\\
\textsuperscript{2}Tencent AI Lab \textsuperscript{3}WeChat AI   \\
 \texttt{\{lucas.tingchenfu,thisisjcykcd,lemaoliu\}@gmail.com} \\
\texttt{ruiyan@ruc.edu.cn} \\
}

\begin{document}
\maketitle
\renewcommand{\thefootnote}{\fnsymbol{footnote}}
\footnotetext[2]{This work was done during internship at Tencent AI Lab.}
\footnotetext[1]{Corresponding authors: Deng Cai and Rui Yan.}
\setcounter{footnote}{0}
\renewcommand{\thefootnote}{\arabic{footnote}}

\begin{abstract}
Supervised fine-tuning (SFT) on instruction-following corpus is a crucial approach toward the alignment of large language models (LLMs). However, the performance of LLMs on standard knowledge and reasoning benchmarks tends to suffer from deterioration at the latter stage of the SFT process, echoing the phenomenon of alignment tax. Through our pilot study, we put a hypothesis that the data biases are probably one cause behind the phenomenon. To address the issue, we introduce a simple disperse-then-merge framework. To be concrete, we disperse the instruction-following data into portions and train multiple sub-models using different data portions. Then we merge multiple models into a single one via model merging techniques. Despite its simplicity, our framework outperforms various sophisticated methods such as data curation and training regularization on a series of standard knowledge and reasoning benchmarks.\footnote{The code is released at {\scriptsize\url{https://github.com/TingchenFu/ACL24-ExpertFusion}}.}     


\end{abstract}

\section{Introduction}
Trained on trillions of tokens from webpages~\citep{achiam2023gpt4,bai2023qwen,anil2023gemini}, large language models~(LLMs) have demonstrated impressive capacity on obtaining general-purpose representations for various downstream NLP tasks. However, pre-trained language models may not follow human instructions~\citep{ouyang2022training} and produce toxic, hallucinated, or biased content~\citep{sun2024trustllm,huang2023survey,zhang2023sirens}. To address the issue, supervised fine-tuning~\citep{ouyang2022training} on instruction-following data has emerged as one of the \textit{de facto} paradigms~\citep{taori2023alpaca,chiang2023vicuna} for aligning LLMs with human preferences.


\begin{figure}
\centering
\begin{subfigure}{0.45\linewidth}
\includegraphics[width=\linewidth]{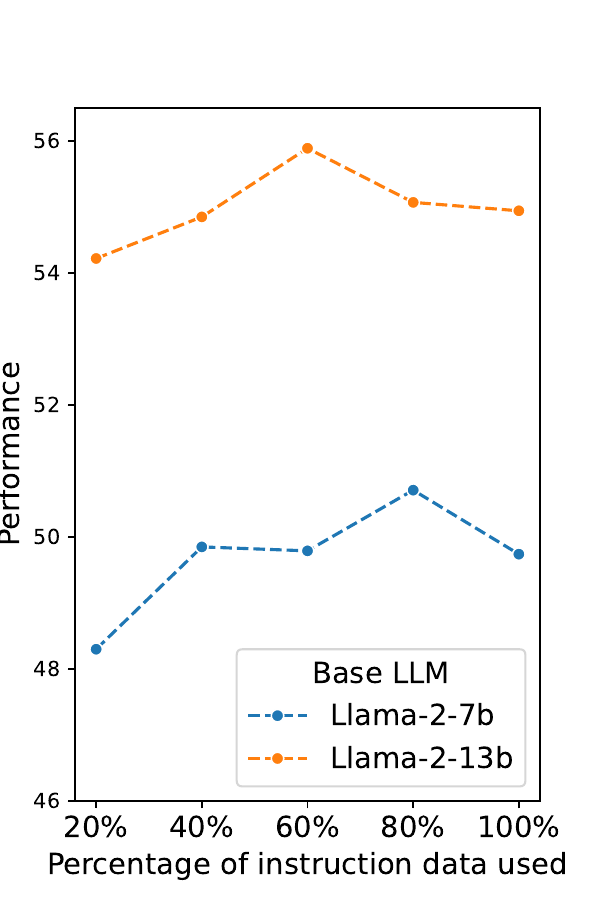}
\caption{The performance on MMLU (5-shot, accuracy) vs. data size.}
\label{fig:tulu-mmlu-updown}
\end{subfigure}
\begin{subfigure}{0.45\linewidth}
\includegraphics[width=\linewidth]{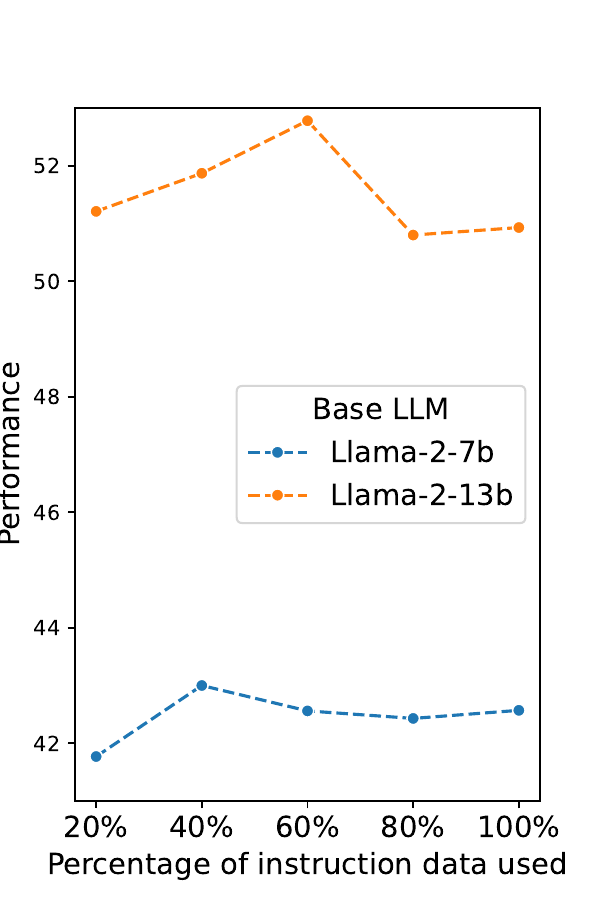}
\caption{The performance on BBH (3-shot, exact match) vs. data size.}
\label{fig:tulu-bbh-updown}
\end{subfigure}
\caption{The performance on MMLU and BBH when tuning Llama-2-7b and Llama-2-13b with different sizes of instruction-following data from \tulu-V2-mix.}
\label{fig:tulu-updown}
\end{figure}

However, with the size of instruction-following data rising, it has been observed that the performance of LLM on standard knowledge and reasoning benchmarks does not always improve but exhibits degradation~\citep{dou2023loramoe}, i.e., the alignment tax~\citep{bai2022training}, as is shown in Figure~\ref{fig:tulu-updown}. In other words, simply scaling up the instruction-following data leads to a quick bump into the upper bound where the marginal return of increasing data size approaches zero or even minus. It is therefore non-trivial to unleash the full potential of large-scale instruction-following data.

Prior studies tend to attribute the alignment tax phenomenon to the low-quality samples within the instruction-following corpus~\cite{chen2023alpagasus,cao2023instruction}, or the knowledge forgetting during the SFT process~\cite{dou2023loramoe,ren2024learning}. However, our pilot study in Section~\ref{sec:pilot} reveals that the quality issue and the catastrophic forgetting of pre-training knowledge are probably not the main cause of the alignment tax since the decline can be observed across corpora with varied sizes and quality. 

By analyzing the trend of loss descent during the SFT process, we alternatively posit that the data biases fitted on the instruction data are probably one of the major causes behind it. Specifically, during the tuning process, LLMs fit on dataset biases while acquiring instruction-following ability. In the beginning, the acquisition of generalizable ability predominates so the performance on knowledge and reasoning benchmarks improves. However, during the tuning process, the learning of generalization quickly stagnates and the model tends to acquire more data biases instead, which harms the parametric knowledge of LLM and leads to a decline in related benchmarks.

We propose a frustratingly simple \modelname~(\textbf{D}isperse-\textbf{T}hen-\textbf{M}erge) framework composed of three steps: (1) we initially distribute the instruction-following data into several clusters and then (2) perform instruction tuning on each cluster of data to obtain a series of sub-models assimilating different data biases; (3) finally we merge the sub-models trained on each cluster into a single one in the weight space, such that the data bias of each sub-model is mitigated at fusion. Importantly, \modelname ensures the reduction of alignment tax when instruction tuning with almost no extra cost at both training and inference.

To empirically verify the efficacy of the \modelname\ framework, we conduct extensive experiments and evaluations across $9$ benchmarks involving math reasoning, world knowledge, and code generation. The experiment results exhibit that \modelname outperforms both (1) data selection methods that filter out low-quality samples~\citep{dou2023loramoe}; and (2) regularization and continue training methods that prevent the forgetting of knowledge learned from pre-training ~\citep{kirkpatrick2016overcoming,polnick2018experience}. In particular, different from previous methods, \modelname does not require any additional training and it incurs almost no extra cost at inference.


The contribution of this paper can be summarized as follows:
\begin{itemize}[wide=0.\parindent,noitemsep,topsep=0.em]
    \item We empirically verify and analyze the effect of alignment tax during the instruction tuning, thereby putting a hypothesis that the dataset biases are the reason behind the alignment tax.
    \item We propose a frustratingly simple \modelname framework in which the biases from instruction-following data are distributed and forgotten.
\end{itemize}
\section{Related Work}
\paragraph{Supervised Instruction Tuning.} 
Supervised fine-tuning of LLMs on open-domain instruction-following data~\citep{ouyang2022training} is a promising approach for calibrating LLMs with human values, which is a critical prerequisite prior to their deployment in real-world scenarios~\citep{xu2023align}. Bypassing the complex and unstable proximal policy optimization algorithm~\citet{schulman2017proximal} in the reinforcement learning from human feedback (RLHF) procedure~\citep{ouyang2022training}, SFT only requires a high-quality instruction-following corpus collected from GPT-4~\citep{achiam2023gpt4} or human annotator~\citep{zhou2023lima,databricks2023dolly2} to tune on. In spite of its simpleness, a surge of recent models~\citep{ding2023enhancing,xu2023wizardlm,geng2023koala,xu2023baize} prove the effectiveness of SFT with their impressive performance on both conventional knowledge and reasoning  benchmarks~\citep{hendrycks2021mmlu} and newly appeared instruction-following benchmarks~\citep{li2023alpacaeval,zheng2023judging}. However, \citet{bai2022training} point out that in particular cases, alignment of LLM is a double-edged sword, enhancing instruction-following ability at the sacrifice of capacity on the conventional knowledge and reasoning benchmark, or the alignment tax. Some follow-ups~\citep{dou2023loramoe,chen2023alpagasus} conjecture that low-quality samples and interference of parametric knowledge are the reasons behind this. Different from previous works, in this study we propose a new perspective to understand the root cause of alignment tax.

\paragraph{Model Merging.} Model merging is an effective technique to aggregate the capacity of multiple models. Distinct from model ensemble, merging techniques involve pruning~\citep{yadav2023tiesmerging}, re-scaling~\citep{yu2023language}, re-weighting~\citep{matena2022merging} or rotating~\citep{singh2020model} the parameters of multiple models before merging them into a single one in the weight space, therefore incurring no extra latency at inference. 
Different from previous works that apply model merging for multi-task learning~\citep{yang2023adamerge,jin2023dataless}, machine unlearning~\citep{hu2023separate,daheim2023model}, domain transfer~\citep{ilharco2023editing,zhang2023composing}, multi-objective reinforcement learning~\citep{rame2023rewarded,jang2023personalized}, we utilize model merging for the alignment of LLM. 
Actually, model merging is closely related to the learned biases of neural networks. Although small models with different prediction mechanisms can hardly be merged together without performance loss~\citep{lubana2022mechanistic,juneja2023linear}, it is different for large-scale fine-tuned models from pre-trained checkpoints, which could generally maintain their capacity when merged together~\citep{qin2022exploring, gueta2023knowledge}. Recently, \citet{zaman2023fuse} and \citet{wan2024knowledge} have shown the possibility of fusing complementary knowledge or removing unintentional memory with the assistance of model merging.

\section{Pilot Study}
\label{sec:pilot}
\begin{figure}
    \centering
    \includegraphics[width=1.0\linewidth]{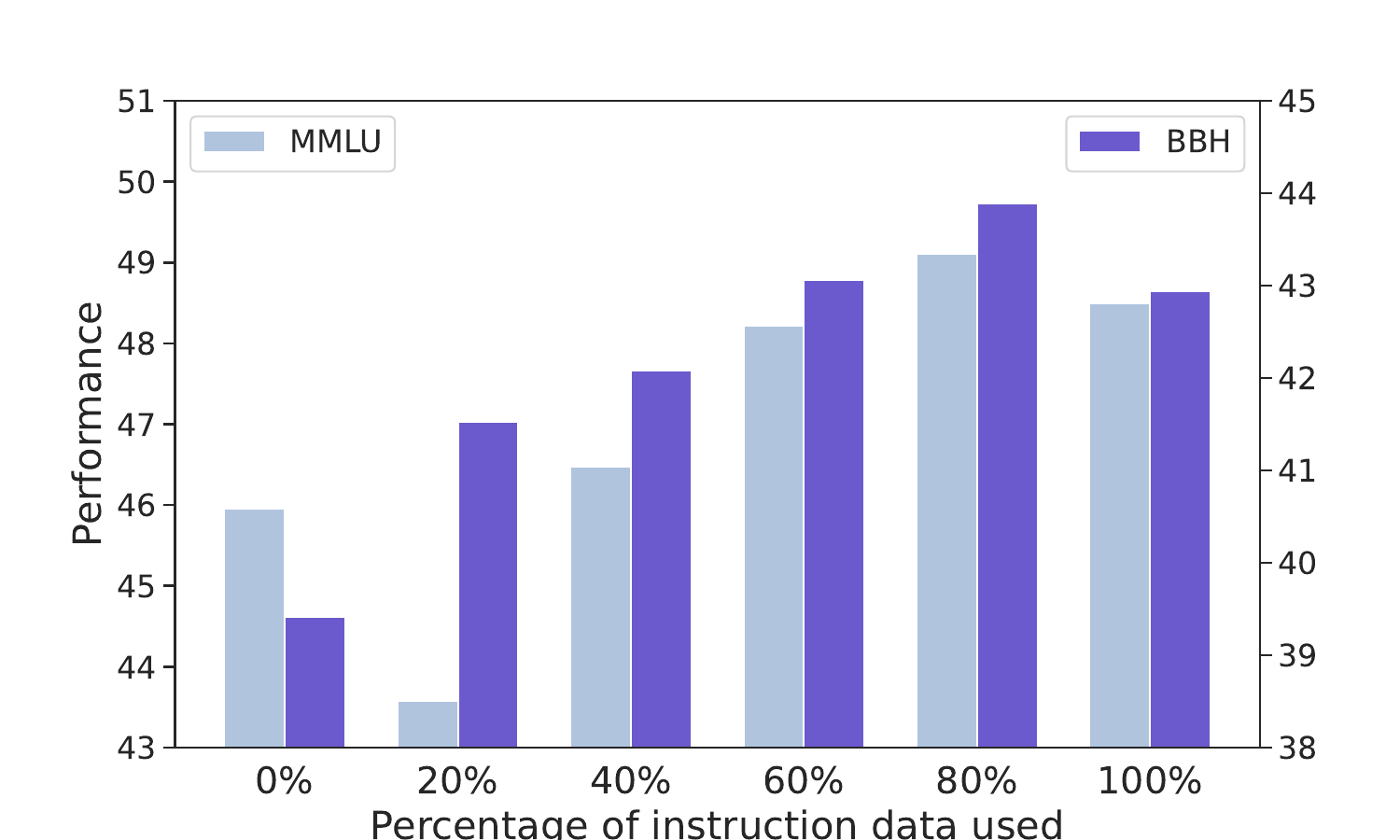}
    \caption{The performance on MMLU (5-shot, accu-
racy) and BBH (3-shot, exact match) when tuning Llama-2-7b with different sizes of instruction-following data from the high-quality subset of \tulu-V2-mix.}
    \label{fig:tulu-above2.5-updown}
\end{figure}

The existence of alignment tax indicates an upper bound of performance when directly increasing the data size at supervised fine-tuning. It is thus necessary to analyze the underlying cause for the alignment tax to unleash the full potential of instruction-following data. Specifically, we first examine the intuitions that data quality and knowledge forgetting are responsible for the decline in conventional knowledge and reasoning benchmarks (Section~\ref{sec:two-reasons}), and then posit our hypothesis that the biases during fitting the instruction-following data is probably one of the major causes (Section~\ref{sec:real-reason}).
\subsection{Are Data Quality and Knowledge Forgetting the Main Causes of the Alignment Tax?}
\label{sec:two-reasons}
\begin{figure}
    \centering
    \includegraphics[width=1.0\linewidth]{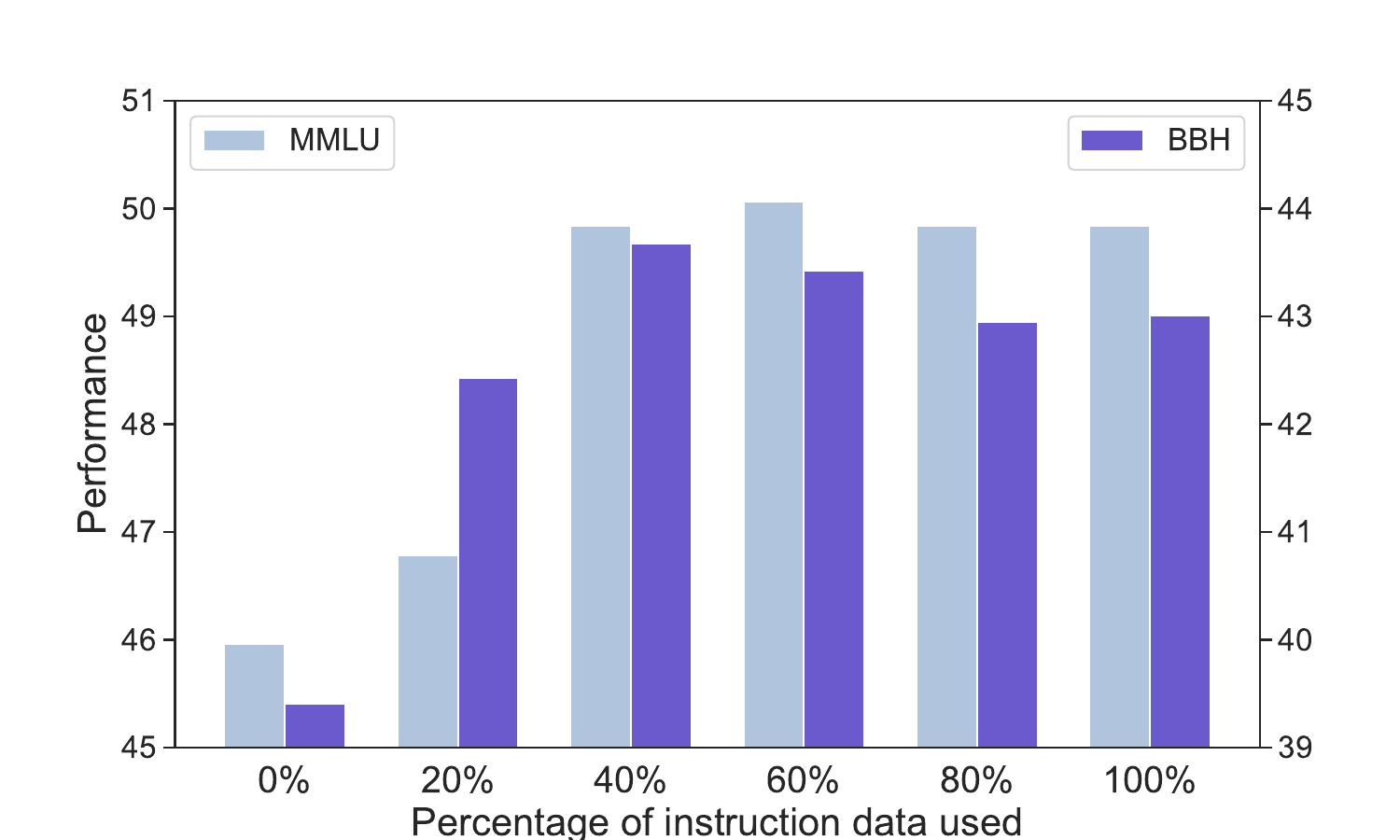}
    \caption{The performance on MMLU (5-shot, accu-
racy) and BBH (3-shot, exact match) when tuning Llama-2-7b with different sizes of instruction-following data from \tulu-V2-mix with Replay.}
    \label{fig:tulu-continue-updown}
\end{figure}

The experiments are mainly conducted on Llama-2-7b~\cite{touvron2023llama2} with \tulu-V2-mix. 
To examine the previously accepted data quality hypothesis~\citep{chen2023alpagasus}, we employ the quality evaluator in \citet{liu2024what} to filter \tulu-V2-mix samples, keeping only the samples with an above 2.5 quality score for tuning and the experiment results are shown in Figure~\ref{fig:tulu-above2.5-updown}. Besides, to verify the effect of pre-training knowledge forgetting, we mix the instruction-following corpus with an equivalent amount of pre-training data from Redpajama~\citep{together2023redpajama} for multi-tasking, and the experiment results are shown in Figrue~\ref{fig:tulu-continue-updown}.


From the experiment results, it is not challenging to discern the following points: 
\begin{itemize}[wide=0.\parindent,noitemsep,topsep=0.em]
\item \textbf{Data quality is probably NOT the main reason}.
Even if we filter out the low-quality samples within the instruction-following corpus with a quality evaluator~\citep{liu2024what}, the alignment tax still exists as shown in Figure~\ref{fig:tulu-above2.5-updown}, suggesting that data quality is probably not the main cause behind the performance decline. 

\item \textbf{Knowledge forgetting is probably NOT the main reason}. Although a significant amount of pre-training data is mixed into the pre-training corpus to alleviate the forgetting and intervention of parametric knowledge, from Figure~\ref{fig:tulu-continue-updown} we can see the drop in performance of traditional knowledge and reasoning benchmarks can hardly be removed. Therefore, it is probably unreasonable to attribute alignment tax to knowledge forgetting.


\end{itemize}
\subsection{Seek for Main Causes of the Alignment Tax}
\label{sec:real-reason}

\begin{figure}
    \centering
    \includegraphics[width=1.0\linewidth]{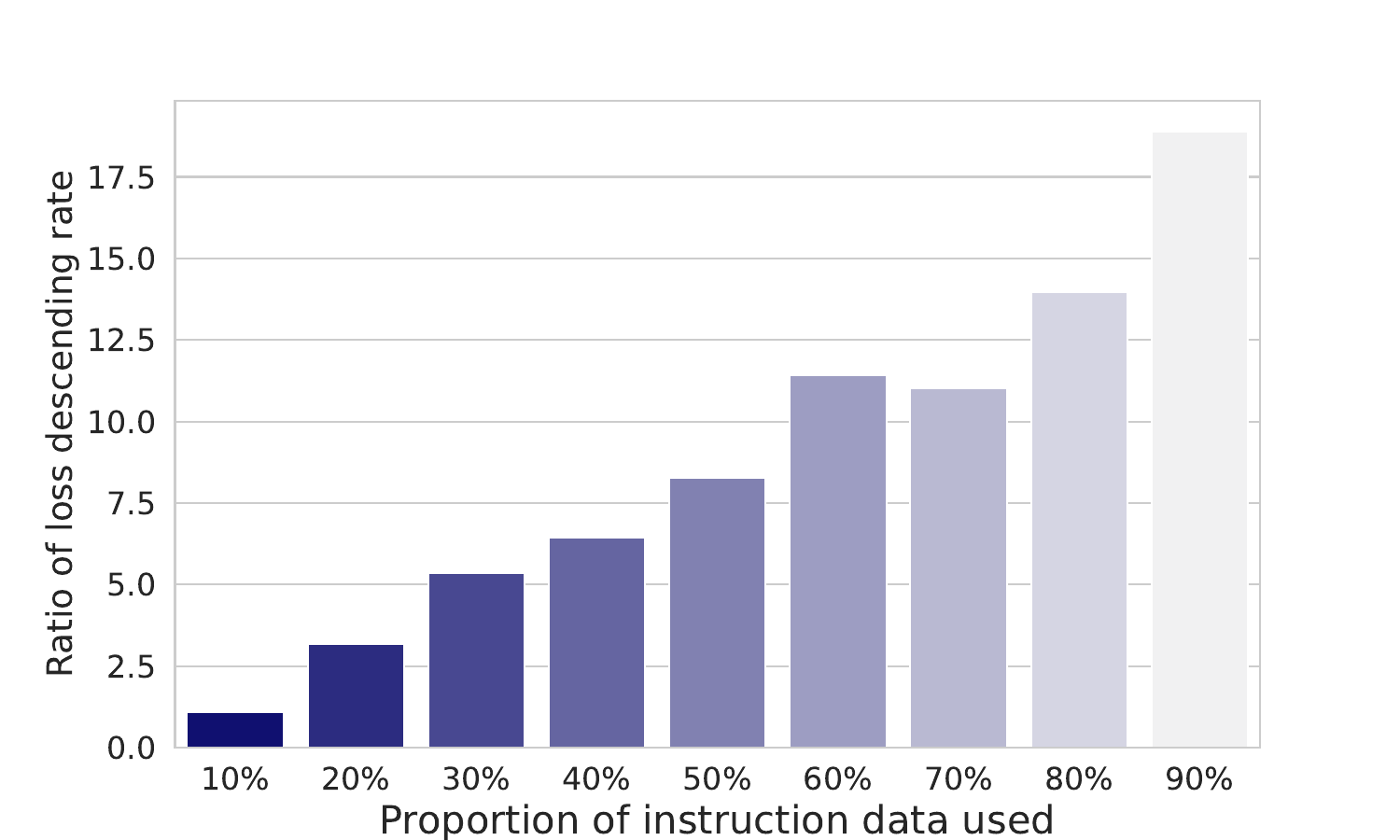}
    \caption{The loss variation ratio between the training set and the validation set, or $\Delta\mathcal{L}_{train}/\Delta\mathcal{L}_{val}$ when tuning Llama-2-7b-hf on \tulu-V2-mix data.}
    \label{fig:loss}
\end{figure}

To understand the reason behind the alignment tax and in particular what is learned when alignment tax occurs, we propose to track the change of loss during the SFT process. In detail, we randomly split the dataset into $10$ portions with equal sizes, training on $9$ of them sequentially and leaving one for evaluation. Every after a portion is finished, we measure the loss reduction on the training set $\Delta\mathcal{L}_{train}$ and the loss reduction on the validation set $\Delta\mathcal{L}_{val}$.  Intuitively, while $\Delta\mathcal{L}_{val}$ reflects the enhancement in generalizable model capacity on instruction following, $\Delta\mathcal{L}_{train}$ encompasses not only the generalizable instruction-following ability, but also the ungeneralizable data specific biases. To measure the composing proportion of the two components, we plot the ratio $\Delta\mathcal{L}_{train}/\Delta\mathcal{L}_{val}$ during the training process in Figure~\ref{fig:loss}.

\begin{figure}
    \centering
    \includegraphics[width=1.0\linewidth]{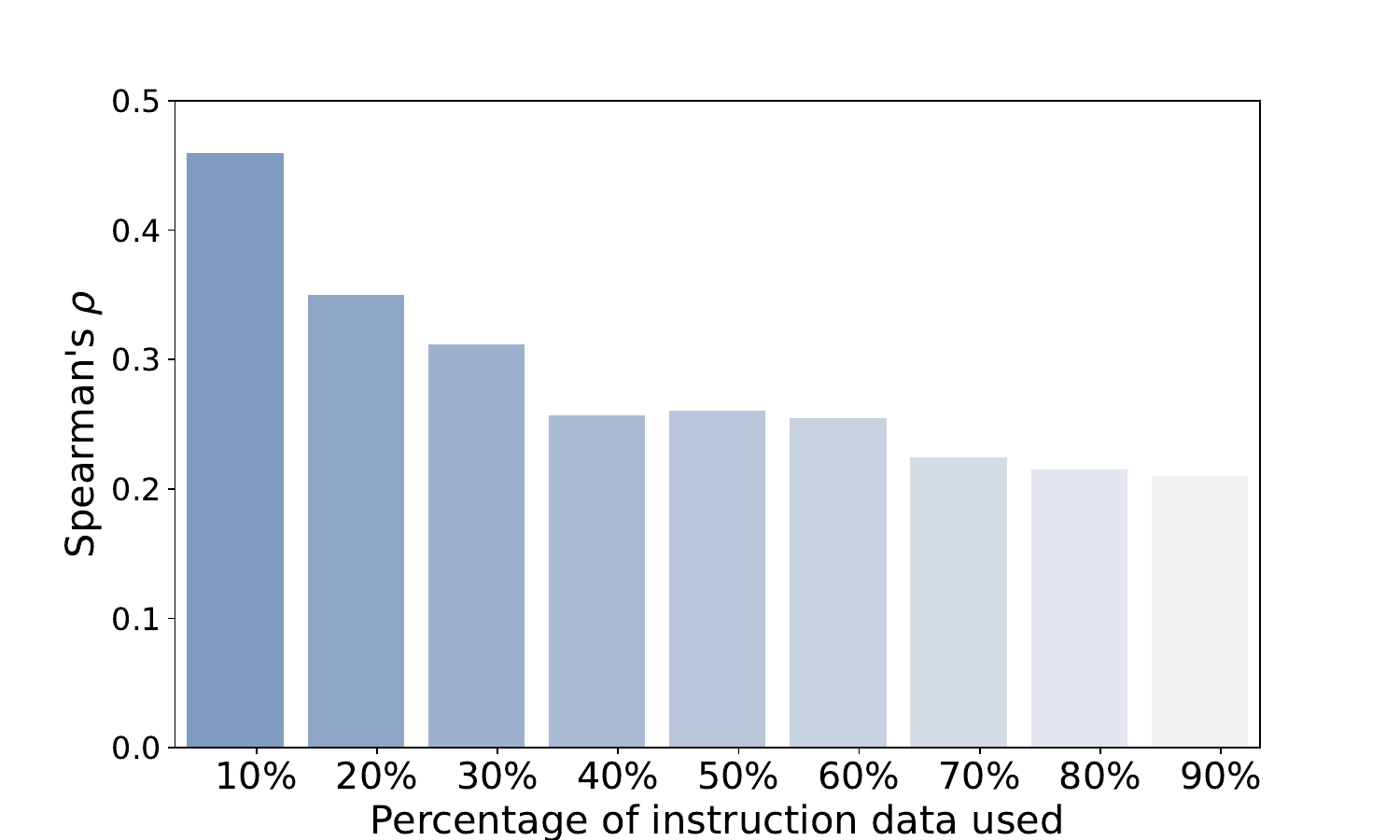}
    \caption{The correlation between training loss reduction and validation loss reduction at token level. }
    \label{fig:corrrelation}
\end{figure}

\begin{figure}
\centering
\begin{subfigure}{0.45\linewidth}
\includegraphics[width=0.9\linewidth]{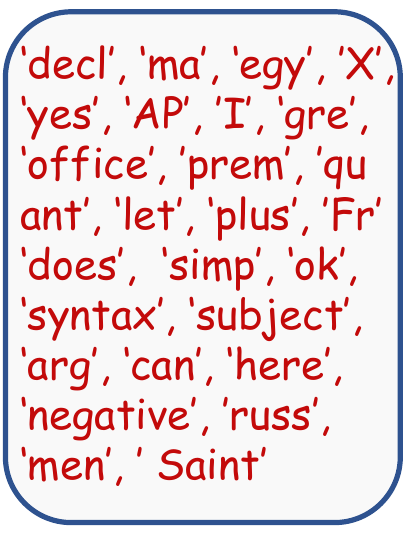}
\caption{Representative loss-reduced tokens after tuned on $10\%$ of instruction-following data.}
\label{fig:loss-word1}
\end{subfigure}
\begin{subfigure}{0.45\linewidth}
\includegraphics[width=0.9\linewidth]{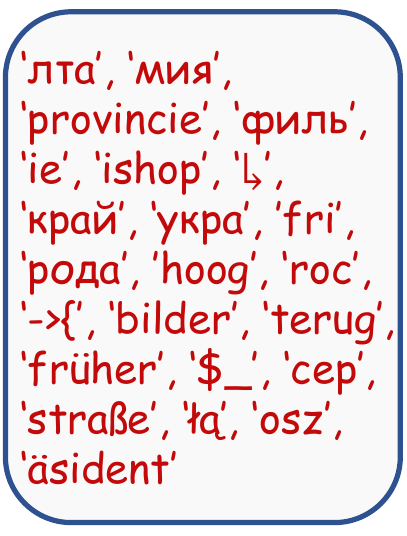}
\caption{Representative loss-reduced tokens after tuned on $90\%$ of instruction-following data.}
\label{fig:loss-word2}
\end{subfigure}
\caption{Representative tokens with prominent loss reduction at different periods of instruction tuning. }
\label{fig:loss-word}
\end{figure}

As is shown, the ratio is approximately $1.0$ at the beginning, suggesting that generalizable instruction-following ability dominates at the initial of training. But as the SFT goes on, the ratio quickly inflates from $1.0$ to nearly $20$, indicating that the acquisition of data biases gradually outweighs other factors and becomes the major reason for loss reduction.  
Furthermore, to have a more intuitive understanding of data-specific biases, we exhibit the token-level biases by measuring the correlation between the per-token loss reduction on the training set and the validation set. Spearman's $\rho$ between the loss reduction on two sets is shown in Figure~\ref{fig:corrrelation}.   
From the figure, it becomes apparent that as the instruction tuning goes on, the fitting on training tokens gradually deviates from the generalizable ability. Meanwhile, some representative tokens with prominent loss reduction at the beginning and the end of training are shown in Figure~\ref{fig:loss-word}. In a comparison between Figure~\ref{fig:loss-word1} and Figure~\ref{fig:loss-word2}, we can observe that the training loss reduction at the end can be mainly attributed to rare words and symbols, suggesting the existence of ungeneralizable data biases.  

Therefore, we hypothesize that the dataset-specific biases and shortcuts~\citep{wang2022identifying,du2021towards} are probably one of the primary contributors to the fitting of the training corpus. Once the assimilation of ungeneralizable dataset biases outweighs the growth of instruction-following capacity, the world knowledge and commonsense reasoning ability of LLM is damaged, thus causing the degradation in related benchmarks, or the alignment tax~\citep{bai2022training}.

\section{Methodology}

\begin{figure*}
    \centering
    \includegraphics[width=0.9\linewidth]{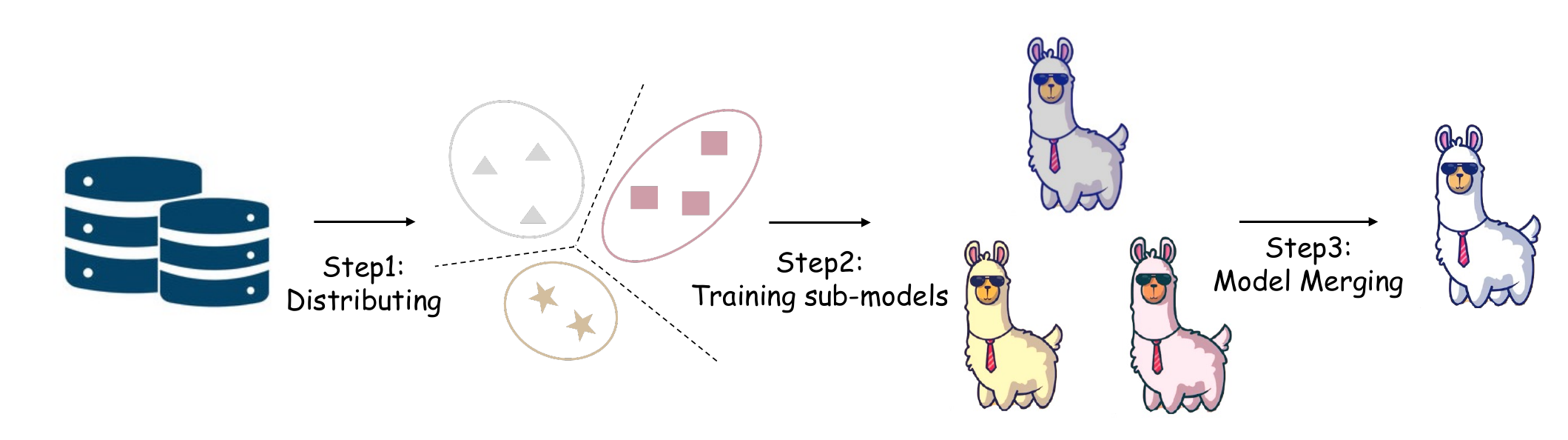}
    \caption{The workflow of \modelname framework, which is composed of three steps: instruction-following data distributing, sub-model training and the merging of sub-models to obtain the final instruction-tuned model.}
    \label{fig:workflow}
\end{figure*}


As analyzed above, vanilla SFT on the full volume of instruction-following data suffers from the assimilation of dataset biases, leading to inefficiency in exploiting large-scale instruction-following corpus. 
Previously, \citet{zaman2023fuse} discovered that when two BERT-based classification models are merged together, the unshared knowledge within each model is mostly forgotten while the common knowledge is enhanced. Getting inspiration from this, to unleash the full potential of large-scale instruction-following data, 
we propose a \modelname framework, as shown in Figure~\ref{fig:workflow}. In a nutshell, we disperse the instruction-following data into multiple portions to obtain a series of sub-models with different data biases. Then through the fusion of multiple sub-models, we can aggregate their instruction-following capacities and eliminate their dataset biases at the same time.


\subsection{Instruction-following Data Distributing}
As standard SFT, \modelname assumes access to a base LLM $\mathcal{M}_0$ and an instruction-following corpus $\mathcal{D}=\{(x_i,y_i)\}_{i=1}^N$ with $N$ samples where $x_i$ is the instruction prompt and $y_i$ is the response. The first step involves distributing the samples into non-overlapped $K$ clusters $\{\mathcal{D}_1,\mathcal{D}_2, \ldots, \mathcal{D}_K\}$. Numerous approaches can be employed to achieve the clustering of the training data. For instance, we can first obtain the embedding of an instruction sample exploiting an off-the-shelf sentence embedding model, or feed the sample into an LLM and use the pooling of the last hidden states as the sentence embedding alternatively. Once the embedding of instruction is obtained, K-means clustering based on cosine distance in embedding space is a good choice to divide the instruction-following corpus into $K$ portions while other clustering schemes like random splitting are also acceptable. 

\subsection{Sub-model Training}
After data distributing, the base LLM $\mathcal{M}_0$ is tuned on $K$ portions of instruction-following data respectively with the standard next-token prediction objective, resulting in $K$ instruction-tuned sub-models $\mathcal{M}_1, \mathcal{M}_2, \ldots,  \mathcal{M}_K$. 
It is worth mentioning that the sub-models are not impervious to biases; however, the biases they acquire vary from one another. In addition, according to the observed trend in Section~\ref{sec:pilot}, fitting on bias diminishes when the scale of instruction-following data narrows down, suggesting that the bias learned by $\mathcal{M}_1, \mathcal{M}_2, \ldots,  \mathcal{M}_K$ is less than the vanilla SFT counterpart.

\subsection{Model Merging}
\label{sec:merge}
The $K$ tuned model $\mathcal{M}_1, \mathcal{M}_2, \ldots,  \mathcal{M}_K$ share common knowledge and capacity on instruction following but their data biases are distinct from each other. Consequently, the acquired capacity on instruction following is maintained, while their unique data biases are forgotten if we fuse the $K$ sub-models together, according to \citet{zaman2023fuse}. Various methods can be utilized to accomplish model fusion and simply taking the weighted average of $K$ sub-models is the most straightforward strategy: 
\begin{equation}
    \mathcal{M}_{f}^i = \sum_{j=1}^K \alpha_j \mathcal{M}_j^i,
\end{equation}
where $\mathcal{M}_f$ is the fused model and the superscript denotes a single parameter in the model.  $\alpha_j$ is the merging weight of the $j$-th sub-model $\mathcal{M}_j$ and we have $\sum_{j=1}^K \alpha_j=1, \alpha_j \geq 0 \ (j=1,2,\ldots,K) $. If not specified otherwise, we use the weighted average of sub-models for its simplicity and ease of use. 

\section{Experiment}
\subsection{Experiment Setup}
\paragraph{Data and Backbone} In our experiment, we employ the \tulu-V2-mix~\citep{ivison2023camels} for SFT, a meticulously curated combination on the basis of \tulu-V1-mix~\citep{wang2023how}. It contains $326,154$ samples collected from $11$ open-sourced instruction-following corpora, which are either manually written by human annotators, converted from existing NLP benchmarks, or curated by GPT-4. As for the backbone, we employ the Llama-2-7b~\citep{touvron2023llama2} as our base LLM. The code is released to facilitate future relevant research.


\begin{table*}[]
\centering
\resizebox{0.95\linewidth}{!}{

\begin{tabular}{lcccccccccc}
\toprule
             & GSM8K & MMLU  & BBH    & ARC-c & OBQA & RACE  & HumanEval & MBPP & TruthfulQA \\
\midrule
Llama-2-base & 13.57 & 45.96 & 39.41  & 43.34 & 31.40 & 39.52 & 12.20     & 20.60 & 24.85 \\
\midrule
Vanilla        & 18.50 & 49.74 & 42.78  & \underline{46.93} & 32.80 & 40.57 & \underline{17.68}  & 21.40 & 25.83      \\
L2-norm        & 18.27 & 49.98 & \underline{43.61}	& 46.33	& 32.4	& 39.62	& 16.46	 & \underline{22.60}	 & 27.66      \\
EWC          & 15.77 & 49.02 & 41.80  & \underline{46.93} & 32.40 & 39.43 & 15.85     & 22.40 & \underline{28.52}      \\
Replay     & 18.27 & 49.46 & 43.05  & 46.76 & 32.20 & 40.19 & 15.24     & 22.40 & 26.32      \\
Uniform Soup         & \underline{19.03} & \underline{50.24} & 42.92  & 46.16 & \underline{33.20} & 40.67 & 14.02     & 21.20 & 25.95      \\
MoE          & 14.48 & 47.36 & 40.39  & 44.62 & 32.00 & 40.10 & 13.41     & 21.80 & 26.07      \\
Deita  & 18.12 & 48.50 & 42.90 & 44.79 & 32.00 & \textbf{41.43} & 15.24 & 20.80 & 28.37  \\
\midrule
\modelname (Ours)         & \textbf{20.62} & \textbf{50.43} & \textbf{44.46}  & \textbf{48.72} & \textbf{33.80} & \underline{41.34} & \textbf{18.29}     & \textbf{23.60} & \textbf{29.13}     \\
\bottomrule
\end{tabular}
}
\caption{Evaluation performance of our training method and its peers. The numbers in bold are the best results and the numbers underlined are the second-best ones. 
}
\label{tab:main}
\end{table*}

\paragraph{Baseline Method} We compare the proposed \modelname\ framework with the following baselines:

\begin{itemize}[wide=0.\parindent,noitemsep,topsep=0.em]
\item \textbf{Vanilla}, or traditional SFT on the instruction-following data with language modeling objective. 
\item \textbf{L2-norm}, where L2 regularization is incorporated in the training objective to circumvent the overfit on instruction-following data and interference with the parametric knowledge. 
\item \textbf{EWC (Elastic Weight Consolidation)}~\citep{kirkpatrick2016overcoming} is a typical regularization in the subfield of continue learning to alleviate the forgetting of previously learned knowledge. There, we apply EWC in SFT to mitigate the catastrophic forgetting of pre-training knowledge. 
\item \textbf{Replay}~\citep{polnick2018experience} is another typical method for mitigating catastrophic forgetting in continue learning. In our implementation, we mix the pre-training data reconstructed by Redpajama~\citep{together2023redpajama} into the instruction-following corpus in a 1:1 ratio and perform multi-task learning on plain language modeling and instruction-following to retain the pre-training knowledge. 
\item \textbf{Uniform Soup}~\citep{wortsman2022model} is a similar recipe to ours in the sense that it fuses multiple trained models into a single one employing model merging techniques. However, in this case, multiple models are trained on the entire corpus with different hyper-parameter configurations.  
\item \textbf{MoE}~\citep{dou2023loramoe} is a recently proposed method to deal with the performance drop of LLM in knowledge-intensive benchmarks. Combining MoE with parameter-efficient fine-tuning, \citet{dou2023loramoe}  enables expert coordination for task utilization and full leverage of parametric knowledge.  

\item \textbf{Deita}~\citep{liu2023what} is an automatic data selection strategy for alignment comprehensively considering the complexity, quality, and diversity of instruction-following data. In our implementation, we keep the samples with complexity scores exceeding 2.5 for training.

\end{itemize}

\begin{table*}[]
\centering
\resizebox{0.95\linewidth}{!}{

\begin{tabular}{lcccccccccc}
\toprule
             & GSM8K & MMLU  & BBH   & ARC-c & OBQA & RACE  & HumanEval & MBPP & TruthfulQA \\
\midrule
Ours         &\textbf{20.62}	& \textbf{50.43}	 & \textbf{44.46}  &	\textbf{48.72} & 33.80 & 41.34 & \textbf{18.29} & 23.60 & 29.13      \\
\midrule
MiniLM (I+R)      & 16.76 & 50.04 & 42.98  & 47.78 & 32.00 & 41.63 & 15.24     & 23.20 & 30.72      \\
MiniLM (I)  & 19.71	& 49.95	& 42.67	& 47.01	&33.40 & 41.53	& 15.24	& 21.20	& \textbf{30.97}   \\
MiniLM (R) & 18.57	& 49.75	& 43.17	& 47.95	& \textbf{34.20}	& \textbf{42.39}	& 15.85	& \textbf{24.60}	& 29.50\\
\midrule
MPNet (I+R)       & 16.83 & 49.73 & 42.94 & 47.87 & 32.80 & 41.34 & 15.24     & 23.80 & \textbf{30.97}      \\
MPNet (I) & 14.86	& 49.88	& 42.40 & 48.38	&33.40	& 41.53	& 14.63	&20.80	& 29.87\\
MPNet (R) &16.76 & 49.44 & 43.05 & 48.63 & 32.80 & 42.01 &	15.85&	21.80 & 29.62\\
\bottomrule
\end{tabular}

}
\caption{Evaluation performance with different clustering methods. Numbers in bold are the best results.}
\label{tab:cluster}
\end{table*}

\paragraph{Evaluation} To have a comprehensive understanding on the efficacy of different training recipes, the evaluation encompasses the capacity of LLM in multiple aspects: math reasoning (GSM8K~\citealp{cobbe2021training}), factual knowledge (MMLU~\citealp{hendrycks2021mmlu}), commonsense reasoning (BBH~\citealp{suzgun2023bbh}, ARC-c~\citealp{clark2018arc} and OpenBookQA~\citealp{mihaylov2018suit}), reading comprehension (RACE~\citealp{lai2017race}), code generation (HumanEval~\citealp{chen2021codex} and MBPP~\citealp{austin2021program}) and truthfulness (TruthfulQA~\citealp{lin2022truthfulqa}), strictly following the evaluation protocol of Open LLM Leaderboard\footnote{\scriptsize\url{https://huggingface.co/spaces/HuggingFaceH4/open_llm_leaderboard}}. In addition, we assess their instruction following ability with MT-bench~\citep{zheng2023judging} and Vicuna-bench~\citep{chiang2023vicuna}, two widely used instruction-following benchmarks. 

\subsection{Experiment Results}

The main experiment results are shown in Table~\ref{tab:main}, in which we randomly distribute the instruction-following data into $K=4$ clusters and utilize average weight merging ($\alpha_j=0.25, j=1,2,3,4$) for fusion. From the table we can observe that our proposed approach outperforms its peers on most evaluation benchmarks, proving the effectiveness of our \modelname framework. Meanwhile, the performance of Uniform Soup is notable, achieving the second-best results on three benchmarks. The difference between Uniform Soup and ours lies in that their sub-models for merging are trained on the full volume of data with different hyper-parameters. Consequently, the data biases of its sub-models are more likely to be overlapped and cannot be removed at merging. In addition, the performance of L2-norm and EWC also attains impressive performance on two benchmarks respectively, possibly due to the retention of pre-training knowledge through regularization techniques.

\paragraph{The effect of different clustering methods.} To investigate the impact of different data clustering methods, we experiment with different sentence embedding models and different encoding schemes. In detail, we use MiniLM~\citep{wang2020minilm} and MPNet~\citep{song2020mpnet} from the SentenceTransformers library~\citep{reimers2019sentencebert}\footnote{\scriptsize\url{https://www.sbert.net/}} to encode the instruction (I) or the response (R) or both of them (I+R) to obtain their dense representation for K-means clustering. The experiment results based on different clustering methods are shown in Table~\ref{tab:cluster}. From the table, it can be inferred that although the dense representation obtained via encoding response (R) is slightly better than other encoding schemes for clustering, none of those sophisticated clustering methods have an obvious advantage over simple random clustering.

\begin{table}[]
\centering
\resizebox{0.8\linewidth}{!}{
\begin{tabular}{lcc}
\toprule
         & MT-bench & Vicuna-bench \\
\midrule
Vanilla    & 4.86     & 6.26         \\
L2-norm       & 4.61     & 6.39         \\
EWC      & 4.44     & 6.46         \\
Replay & 4.78     & 5.75         \\
Uniform Soup     & \underline{5.04}     & \textbf{7.48}         \\
MoE      & 3.67     & 6.43         \\
Deita  & 4.71 & 6.20   \\
\midrule
\modelname (Ours)     & \textbf{5.19}     &\underline{6.60}         \\
\bottomrule
\end{tabular}
}
\caption{The evaluation results of instruction following ability on MT-bench and Vicuna-bench. the numbers in bold are best results. The numbers underlined are the second-best ones.}
\label{tab:llmbench}
\end{table}

\paragraph{The effect of different merging methods.}
As for the model fusion, we experiment with several widely used merging techniques: (1) \textbf{Fisher}~\citep{matena2022merging} employs the approximated Fisher information matrix to approach the fused model with the highest joint probability. (2) \textbf{Task Vector}~\citep{ilharco2023editing} subtracts the base LLM weight from the instruction-tuned model in the weight space to obtain the task vector and accomplish merging with vector arithmetic; (3) \textbf{Tie Merge}~\citep{yadav2023tiesmerging} trims and prunes the models before merge and resolves the interference between multiple models.   (4) \textbf{DARE}~\citep{yu2023language} refines task vector by dropout and re-scale before conducting vector arithmetic. 
The experiment results on different clustering methods are shown in Table~\ref{tab:merge}. Similarly, it seems that no single merging method is apparently superior to others, and simple average weight merging is sufficient.

\begin{table*}[]
\centering
\resizebox{0.95\linewidth}{!}{

\begin{tabular}{lcccccccccc}
\toprule
             & GSM8K & MMLU  & BBH   & ARC-c & OBQA & RACE  & HumanEval & MBPP & TruthfulQA \\
\midrule
Ours         &\textbf{20.62}	& \textbf{50.43}	 & \textbf{44.46}  &	48.72 & 33.8 & 41.34 & \textbf{18.29} & \textbf{23.60} & \textbf{29.13}      \\
\midrule
Fisher & 19.64 & 50.41 & 44.28 & 48.04 & \textbf{34.40} & 41.53 & 17.68 & 22.40 & 28.52 \\
Task Vector & 19.71 & 49.85 & 43.58  & \textbf{49.66} & 33.40 & 41.82 & 17.68     & 22.40 & 28.27      \\
Tie Merge   & 18.42 & 49.32 & 42.90  & 47.10 & 33.00   & 40.57 & 16.46     & \textbf{23.60} & 27.66      \\
DARE        & 18.95 & 49.89 & 43.37  & 49.15 & 33.40 & \textbf{42.30}  & 16.46     & 22.00   & 28.64     \\
\bottomrule
\end{tabular}

}
\caption{Evaluation Performance with different merging methods. Numbers in bold are the best results among different merging methods.}
\label{tab:merge}
\end{table*}

\begin{figure}
    \centering
    \includegraphics[width=0.95\linewidth]{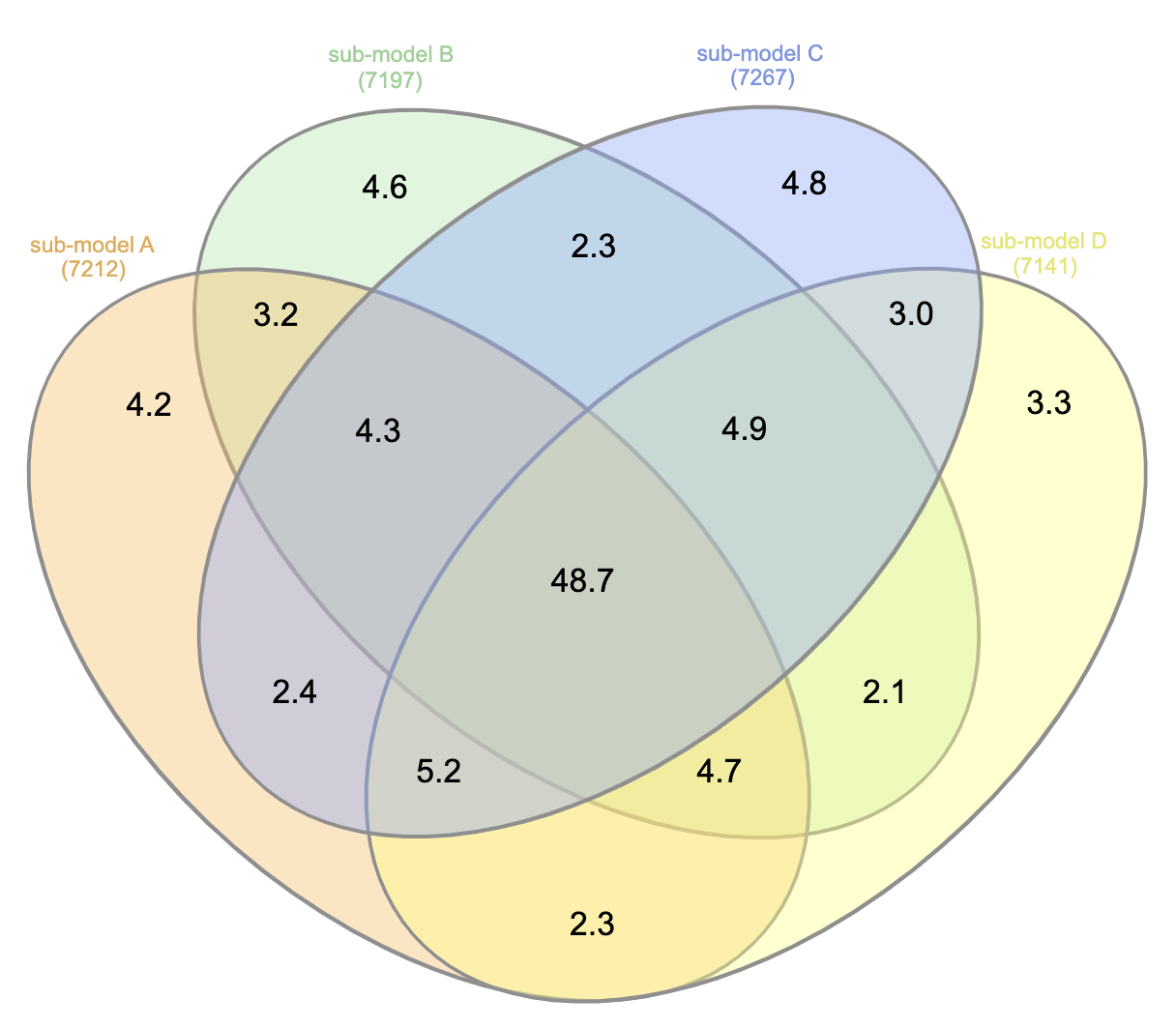}
    \caption{The Venn diagram for the error set of $K=4$ sub-models on MMLU. The numbers denote the percentage of error cases in that particular set relative to all error cases.}
    \label{fig:venn}
\end{figure}

\paragraph{Performance on instruction following.} The experiment results on instruction-following ability are shown in Table~\ref{tab:llmbench}. We can observe that our approach surpasses the vanilla SFT (5.19 v.s. 4.86 in MT-bench and 6.60 v.s. 6.26 in Vicuna-bench) and attains the best or the second-best performance, suggesting that our approach not only maintains the basic knowledge and reasoning ability of language model, but also improve the instruction-following ability. Notably, Uniform Soup exhibits strong instruction-following ability since their sub-models are trained on a full volume of data and therefore acquire stronger instruction-following capacity, although at the cost of more damage to world knowledge and commonsense reasoning ability.


\section{Analysis}
\label{sec:analysis}
The SFT experiments on \tulu-V2-mix have proven the efficacy of the proposed approach. 
To gain more in-depth insights, further exploration and analysis are detailed below.

\begin{figure}
    \centering
    \includegraphics[width=1.0\linewidth]{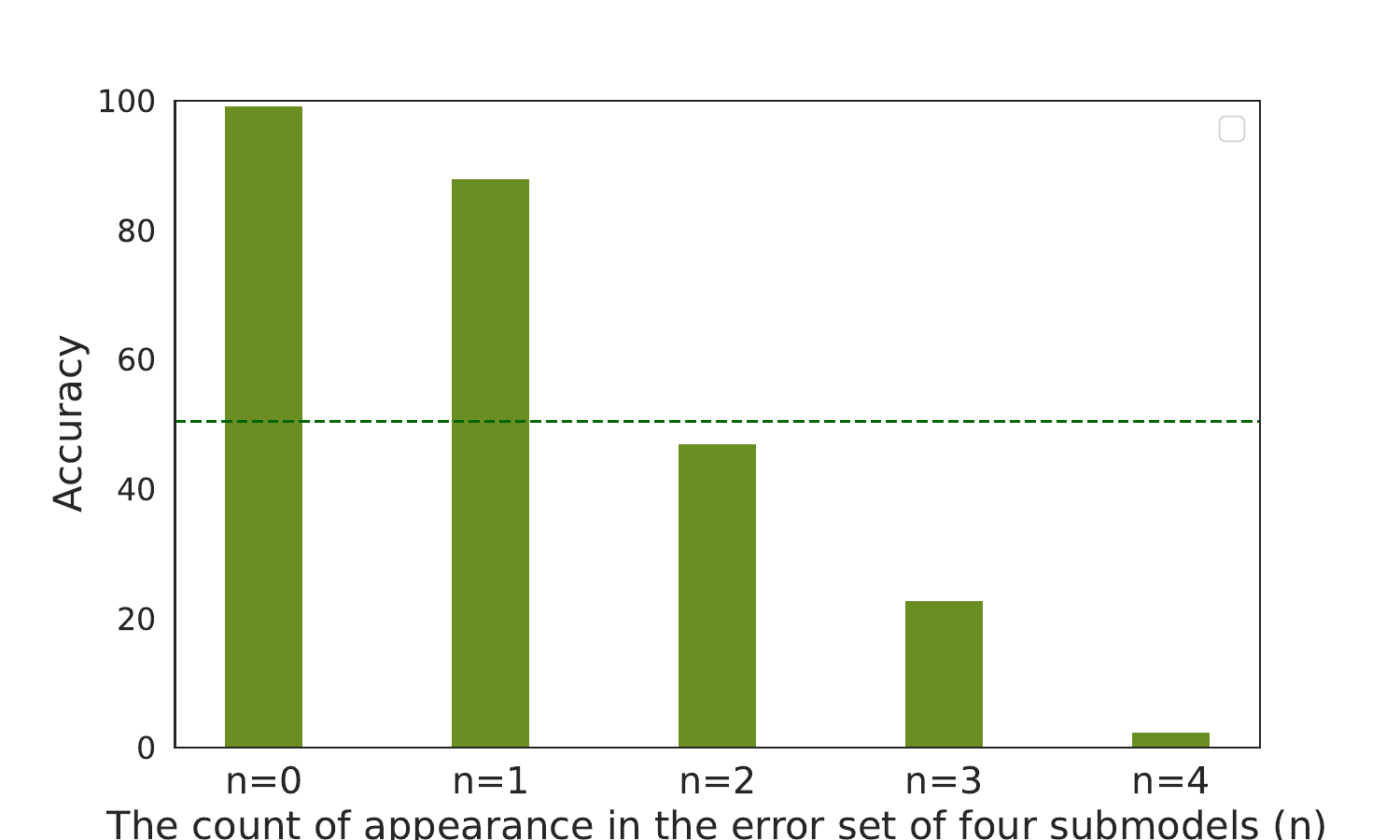}
    \caption{The accuracy of the fused model v.s. the count of appearance in the error sets of sub-models. The dotted line denotes the overall accuracy on MMLU.}
    \label{fig:shortcut}
\end{figure}

\textbf{Question1: How does \modelname help? \\Answer1:} Measuring the data biases or shortcuts on instruction following is challenging since we are agnostic to the specific form of the bias.  Therefore, we choose to quantify the data biases of LLM through their error sets on MMLU~\citep{hendrycks2021mmlu}. We plot the Venn diagram for the error set of $K = 4$ sub-models in Figure~\ref{fig:venn}. It can be observed that their error sets share a large portion (48.7\%) but every sub-model has its own error cases (accounting for $4.2\%, 4.6\%, 4.8\%$ and $3.3\%$ of the entire error cases respectively), attributing to their unique shortcut. 

Next, we bucket the test case of MMLU into different bins according to their count of appearance $n$ in the error sets. For example, $n=4$ for the cases within the common intersection of four error sets while $n=1$ for unique error cases of sub-models. Specifically, $n=0$ denotes that the case does not belong to any error set or equivalently all sub-models can figure out the answer correctly. Then we plot the accuracy of the fused model on each bin in Figure~\ref{fig:shortcut}. From the figure, 
the accuracy on the first bin ($n=0$) is nearly approaching $100\%$, suggesting that the common knowledge is retained. On the other hand, the high accuracy on the second bin means that the unique error cases of four sub-models are likely to be correctly solved by the fused model, which is evidence for the forgetting of unique data biases in four sub-models.

\begin{table}[]
\centering
\resizebox{1.0\linewidth}{!}{
\begin{tabular}{lccccc}
\toprule
\multicolumn{1}{l}{}           &       & MMLU  & BBH   & ARC-e & ARC-c \\
\midrule
\multirow{2}{*}{Alpaca-GPT4}    & Vanilla  & 47.14 & 39.38 & 77.65 & 45.14 \\
                               & \modelname  & 47.60 & 39.99 & 78.28 & 47.27 \\
\midrule
\multirow{2}{*}{Code-Alpaca}    & Vanilla & 47.04 & 39.04 & 77.82 & 45.31 \\
                               & \modelname  & 47.37 & 40.17 & 77.90 & 45.65 \\
\midrule
\multirow{2}{*}{Baize}         & Vanilla & 44.96 & 39.68 & 74.58 & 43.69 \\
                               & \modelname  & 46.24 & 40.18 & 75.38 & 45.73 \\
\midrule
\multirow{2}{*}{Camel}         & Vanilla & 44.72 & 40.44 & 75.25 & 42.58 \\
                               & \modelname  & 45.81 & 41.32 & 75.51 & 44.11 \\
\midrule
\multirow{2}{*}{Evol-Instruct} & Vanilla & 47.19 & 42.40 & 77.82 & 45.90 \\
                               & \modelname  & 47.24 & 41.69 & 78.28 & 47.70 \\
\midrule
\multirow{2}{*}{LIMA} & Vanilla & 46.70& 39.46 & 76.30 & 43.52    \\
                      & \modelname & 46.25 &  40.06 & 76.85 & 44.28     \\
\bottomrule
\end{tabular}
}
\caption{Experiment results after tuning the base LLM on five widely used instruction-following corpora.}
\label{tab:single-dataset}
\end{table}

\begin{table*}[]
    \centering
    \resizebox{0.95\linewidth}{!}{
    

\begin{tabular}{lcccccccccc}
\toprule
             & GSM8K   &  MMLU  & BBH   & ARC-c & OBQA & RACE & HumanEval & MBPP & TruthfulQA  \\
\midrule
Mistral-7b & 39.95 & 62.56 & 56.08	& 50.34 &	\textbf{32.60} &	40.86 & 28.65 & 39.60 & 28.27        \\
Vanilla SFT & 38.51	& 62.01	& 59.64	& 54.10 &	31.80 &	42.49 &	30.49 & 39.80 &	28.15  \\
Ours  & \textbf{43.52} & \textbf{62.63} & \textbf{60.87} & \textbf{55.80} & 32.20 & \textbf{42.87} & \textbf{31.10} & \textbf{41.40} & \textbf{30.11}   \\
\midrule
Baichuan-2-7b & 21.15 & \textbf{54.33} & 34.75 & 42.15 & 30.60 & 38.28 & 18.29 & 24.20 & 23.01 \\
Vanilla SFT  & 25.63 & 52.18 & 40.53 & 41.72 & 28.80 & 39.52 & 18.90 & 23.40 & 25.70 \\
Ours & \textbf{26.46} & 53.92 & \textbf{42.40} & \textbf{43.52} & \textbf{31.00} & \textbf{40.57} & \textbf{23.78} & \textbf{25.60} & \textbf{26.56}  \\
\bottomrule
\end{tabular}

    }
    \caption{Evaluation Performance with different backbones. Numbers in bold are the best results in the block.}
    \label{tab:backbone}
\end{table*}

\textbf{Question2: Does \modelname yield effective results across instruction-following data of varying sizes and domains?} \\ \textbf{Answer:} To examine the robustness and generality of our approach, aside from \tulu-V2-mix, we conduct experiments on other five widely used instruction-following corpora within or not within the \tulu-V2-mix, namely GPT4-Alpaca (generic, 52,002 samples, \citealp{peng2023instruction}), Code-Alpaca (code, 20,022 samples, \citealp{sahil2023codealpaca}), Baize (Quora \& StackOverflow \& medicine, 158,183 samples, \citealp{xu2023baize}), Camel (STEM, 109,740 samples, \citealp{li2023camel}), Evol-Instruct (generic, 70,000 samples, \citealp{xu2023wizardlm}) and LIMA (Stack Exchange and Reddit, 1,000 samples). 
The performance of our approach in comparison with vanilla SFT is shown in Table~\ref{tab:single-dataset}, from which we can conclude that \modelname is not constrained by the domain of the instruction-following data, but its superiority is influenced by the data size. 


\textbf{Question3: How is model fusion in comparison to model ensemble? \\Answer: } Different from model fusion which aggregates the parameter of multiple models in the weight space, model ensemble aggregates multiple models by manipulating their logits. To draw a comparison of their effects, we substitute the model merging procedure in Uniform Soup~\citep{wortsman2022model} and our approach with model ensemble, and the evaluation results on MMLU are shown in Table~\ref{tab:ensemble}. From the table, model ensemble is almost on par with model fusion except that model fusion is marginally better than ensemble overall. However, the computation required by model ensemble is $K$ (the number of sub-models) times larger than the model fusion, and thus its throughput is inferior to the model merge.

\textbf{Question4: Does the proposed approach work on other base LLM? \\Answer: } To answer the question, we conduct experiments with Mistral-7b~\citep{jiang2023mistral} and Baichuan-2-7b~\citep{yang2023baichuan2}, two renowned backbones with remarkable performance on Open LLM Leaderboard with similar parameter scale. The experiment results are shown in Table~\ref{tab:backbone}, suggesting that \modelname is agnostic to the base LLM and able to generalize to more capable LLMs.

\begin{table}[]
    \centering
    \resizebox{1.0\linewidth}{!}{
    \begin{tabular}{lccccc}
\toprule
              & Humanities & \makecell[c]{Social\\ Science} & STEM & Others & Overall \\
\midrule 
\textit{Uniform Soup}\\
\ Merge    & 47.21           & \textbf{57.36}               & \textbf{40.29}     & \textbf{57.13}      & \textbf{50.24}        \\
\ Ensemble & \textbf{47.31}           & 57.17               & 39.63     & 56.76      & 50.00        \\
\midrule
\textit{Ours} \\
\ Merge    & 47.52           & \textbf{58.27}               & 39.23     & 57.62      & \textbf{50.43}        \\
\ Ensemble & \textbf{47.89}           & 57.20               & \textbf{39.30}     & \textbf{57.90}      & 50.39        \\
\bottomrule
\end{tabular}
    }
    \caption{The evaluation results of model merge and model ensemble on MMLU. Numbers in bold are the best performance in the block. }
    \label{tab:ensemble}
\end{table}

\section{Conclusion}
In this study, we target the alignment tax during the SFT. Through a series of pilot studies, we hypothesize that data biases are the root cause for the decline in standard benchmarks after an LLM goes through the SFT process. To deal with the problem, we propose a simple three-step framework to disperse the biases apart and employ model merging techniques to mitigate the effect of data biases. Extensive experiments are conducted to empirically verify the efficacy of our approach and we hope our research will inspire more future work exploring the essence and mechanism of LLM alignment together with its effects on the capacity of LLM.

\section*{Limitations}

The limitations of this study can be summarized as below:
\begin{itemize}[wide=0.\parindent,noitemsep,topsep=0.em]
    \item In this work, we mainly focus on the alignment tax during the supervised fine-tuning process. Aside from SFT, there are multiple alternative approaches towards the alignment of LLM such as RRHF~\citep{yuan2023rrhf}, DPO~\citep{rafailov2023direct}, and their variants. However, we did not verify or discuss the alignment tax in other alignment methods and we would like to leave this for future work.
    \item We generally utilize LoRA~\citep{hu2022lora} as a parameter-efficient fine-tuning (PEFT) technique for SFT and do not perform experiments with other PEFT techniques such as adapter~\citep{houlsby2019parameter} or IA3~\citep{liu2022few} or full-parameter fine-tuning.
\end{itemize}

\section*{Ethical Consideration}
This paper has few ethical risks and will not pose a problem with ethics.  First, the alignment of large language models is not a new task in natural language processing, and several papers about this task have been published at NLP conferences. Second, all the datasets and benchmarks used in this paper have been published in previous papers. Our work aims at better understanding and eliminating alignment tax towards the tax-free alignment and our approach should not be used for any malicious purpose.

\bibliography{custom}

\appendix
\clearpage
\section{More Details on Experiment Setup}
\subsection{More details on Instruction data}

\begin{table*}[]
    \centering
    \resizebox{1.0\linewidth}{!}{
    \begin{tabular}{llccc}
\toprule
Datasets           & Source                                     & \# Samples & $\Bar{L}_{inst}$ & $\Bar{L}_{output}$ \\
\midrule
FLAN v2~\cite{longpre2023flan}            & NLP datasets + Human-written Instructions  & 49,123      & 327.85       & 15.25     \\
CoT~\cite{wei2022chain}                & NLP datasets + Human-written CoTs          & 49,747      & 151.67       & 32.77     \\
Open Aissatnt 1~\cite{kopf2023ppenassistant}    & Human-written from scratch                 & 7,331       & 20.26        & 149.39    \\
ShareGPT~\cite{chiang2023vicuna}           & User prompts + outputs from various models & 111,912     & 81.09        & 197.71    \\
GPT4-Alpaca~\cite{peng2023instruction}        & Generated w/ Davinci-003 + GPT4            & 19,906      & 16.41        & 107.50    \\
Code-Alpaca~\cite{sahil2023codealpaca}        & Generated w/ Davinci-003                   & 20,016      & 20.81        & 44.94     \\
LIMA$^\star$~\cite{zhou2023lima}               & Human-written from scratch                 & 1,018       & 39.40        & 430.17    \\
Evol-Instruct V2$^\star$~\cite{xu2023wizardlm}   & Generated w/ Davinci-003 + GPT3.5-turbo    & 29,810      & 98.42        & 276.50    \\
Open-Orca$^\star$~\cite{lian2023openorca}          & NLP datasets + GPT-4 generated CoTs        & 29,683      & 154.57       & 110.64    \\
Science literature$^\star$~\cite{dasigi2021dataset} & NLP datasets + Human-written CoTs          & 7,468       & 1118.43      & 45.03     \\
Hardcoded$^\star$          & Human-written from scratch                 & 140        & 5.29         & 69.71    \\
\bottomrule
\end{tabular}
    }
    \caption{The statistics and composition of \tulu-V2-mix. We report the average length of instruction ($\Bar{L}_{inst}$) and the average length of response ($\Bar{L}_{output}$). The datasets marked with asterisk are newly added ones that do not exist in \tulu-V1-mix.}
    \label{tab:statistic-tulu}
\end{table*}

\begin{table*}[]
    \centering
    \resizebox{1.0\linewidth}{!}{
    
\begin{tabular}{llccc}
\toprule
Datasets            & Source                                  & \# Samples & $\Bar{L}_{inst}$ & $\Bar{L}_{output}$ \\
\midrule
Evol-Instruct-70k~\cite{xu2023wizardlm}   & Generated w/ Davinci-003 + GPT3.5-turbo & 70,000      & 77.82        & 206.55    \\
Baize.medical~\cite{xu2023baize}       & Generated w/ ChatGPT                    & 46,863      & 12.41        & 36.13     \\
Baize.quora~\cite{xu2023baize}         & Generated w/ ChatGPT                    & 54,282      & 15.43        & 31.91     \\
Baize.stackoverflow~\cite{xu2023baize} & Generated w/ ChatGPT                    & 57,038      & 19.18        & 26.79     \\
Camel.math~\cite{li2023camel}          & Generated w/ GPT3.5-turbo               & 49,765      & 45.59        & 223.70     \\
Camel.physics~\cite{li2023camel}       & Generated w/ GPT3.5-turbo               & 20,000      & 36.47        & 357.60    \\
Camel.chemistry~\cite{li2023camel}     & Generated w/ GPT3.5-turbo               & 19,983      & 30.94        & 309.20    \\
Camel.biology~\cite{li2023camel}       & Generated w/ GPT3.5-turbo               & 19,992      & 23.89        & 407.51   \\
\bottomrule
\end{tabular}

    }
    \caption{The statistics of other corpus used for tuning at Question 3. We report the average length of instruction ($\Bar{L}_{inst}$) and the average length of response ($\Bar{L}_{output}$).}
    \label{tab:statistic-other}
\end{table*}

In our experiment, we majorly employ the meticulously curated \tulu-V2-mix\footnote{\scriptsize\url{https://huggingface.co/datasets/allenai/tulu-v2-sft-mixture}}~\citep{li2023camel} corpus for SFT. Composed of $11$ subsets, \tulu-V2-mix includes $326,154$ samples, compared to $490,445$ in the V1 mixture. To reduce the computation cost required for fine-tuning, we only keep the first turn of dialogue in case there are multiple instructions and responses in a datum, and the data statistics for each subset consisting of the corpus are shown in Table~\ref{tab:statistic-tulu}. Besides, in Section 6, we perform SFT on other $5$ corpora to investigate the generality of our approach. Among the five corpora, Alpaca-GPT4~\citep{peng2023instruction} and Code-Alpaca~\citep{sahil2023codealpaca} are constituting components of \tulu-V2-mix, while Baize~\citep{xu2023baize}, Camel~\citep{li2023camel} and Evol-Instruct-70k~\citep{xu2023wizardlm} are external instruction corpora and their statistics are shown in Table~\ref{tab:statistic-other}.

\subsection{More details on Evaluation Benchmarks} In our experiment, to draw a comparison with our proposed \modelname framework with other baseline methods, we evaluate on the following benchmarks:

\begin{itemize}
    \item \textbf{GSM8K}~\citep{cobbe2021training} is a collection of 1,319 middle-school level math word problems with each question consisting of basic arithmetic operations ~\citep{azerbayev2023llemma}. Following~\citep{achiam2023gpt4}, we experiment with 8-shot prompting and greedy decoding at inference.  
    \item \textbf{MMLU}~\citep{hendrycks2021mmlu} is a popular aggregated benchmark covering $57$ tasks including elementary mathematics, US history, computer science, law, and more, which are categorized into $4$ subsets: STEM, Humanities, Social Science and Others.   
    Extensive world knowledge and problem-solving ability are required to attain a high score on this benchmark. We use 5-shot prompting at evaluation and report the overall accuracy. 
    \item \textbf{BBH (BIG-Bench Hard)}~\citep{suzgun2023bbh} is a challenging subset of BIG-Bench~\citep{srivastava2022beyond} on which prior language models fall behind average human-raters. Composed of $23$ particularly challenging tasks ($27$ sub-tasks), the benchmark mainly focuses on LLM abilities in four aspects, namely multi-step arithmetic reasoning, natural language understanding, use of world knowledge, and multilingual knowledge and reasoning. Following \citet{suzgun2023bbh}, we evaluate all models via greedy decoding and report the exact match between the generated output (after extracting the content behind the ``the answer is'' keyword) and the ground-truth label. 3-shot prompting and chain-of-thought prompting are employed as a common practice to improve performance. 
    \item \textbf{ARC}~\citep{clark2018arc} is a collection of genuine grad-school level science multiple-choice problems with two subsets, namely Easy Set (ARC-e) and Challenge Set (ARC-c). Our experiments are mainly conducted on ARC-c, which is composed of 1,172 problems that cannot be trivially solved by word co-occurrence algorithm or retrieval algorithm. We adopt zero-shot prompting and report the accuracy.
    
    \item \textbf{OBQA (OpenBookQA)}~\citep{mihaylov2018suit} is a set of elementary level science multiple-choice problems. Modeled after the open book exams testing the understanding of a student on a specific subject, each question in the dataset is accompanied by a basic scientific fact and requires the possession of commonsense knowledge to combine the facts. Similarly,  we adopt zero-shot prompting at inference and report the accuracy.

    \item \textbf{RACE}~\citep{lai2017race} is a large-scale reading comprehension benchmark, in which the problems are collected from the English exams for middle and high school Chinese students and cover a wide range of topics. Compared with other reading comprehension datasets, it requires more reasoning to work out the answer. Similar to the above two benchmarks, we adopt zero-shot prompting at inference and report the accuracy.

    \item \textbf{HumanEval}~\citep{chen2021evaluating} is a suit of 164 hand-written Python programming problems released by OpenAI, with each problem consisting of function signature, docstring, body, and several unit tests to validate the code produced by a language model. Following~\citep{li2023starcoder,rozire2023codellama}, we use $pass@k$ as our metric, which is the total fraction of benchmark problems solved, where a problem is considered solved if any one of $k$ code samples passes every test case. We adopt the simplest version of $pass@k$, namely $pass@1$, which is the likelihood that a problem is solved in a single attempt by the model. Greedy decoding is used for inference. 

    \item \textbf{MBPP}~\citep{austin2021program} is another widely used test set for evaluating the code generation ability of language models, composed of 974 Python short functions and program textual descriptions. Similar to HumanEval, the performance of MBPP is evaluated by $pass@1$ and greedy decoding is adopted for inference. 

    \item \textbf{TruthfulQA}~\citep{lin2022truthfulqa} is a popular problem set for evaluating the truthfulness of LLM. Composed of 817 spanning 38 categories, it is widely used for benchmarking the hallucination of LLM~\citep{zhang2023alleviating,chuang2024dola,li2023inference}. We use the multiple-choice configuration of the benchmark and report the MC1 score, which is the fraction of benchmark problems where models assign the highest scores to the best answer.

    \item \textbf{Vicuna-bench}~\citep{chiang2023vicuna} is a recent benchmark with GPT-4 as a judge. Containing 80 questions spanning various categories such as roleplay, commonsense, and Fermi problems, it evaluates the instruction following proficiency of LLM.  
    
    \item \textbf{MT-bench}~\citep{zheng2023judging} is another rigorous benchmark for measuring both the conversation ability and instruction-following ability of language models. It contains 80 multi-turn questions across eight subjects: writing, roleplay, extraction, reasoning, mathematics, coding, knowledge I (STEM), and knowledge II (humanities/social science). We report the first-turn score since our instruction tuning only involves a single instruction-response pair.
    
 \end{itemize}

 \begin{table*}[]
    \centering
    \resizebox{0.9\linewidth}{!}{
    \begin{tabular}{lccc}
\toprule
                & Llama-2-7b~\cite{touvron2023llama2} & Mistral-7b~\cite{jiang2023mistral}  & Baichuan-2-7b~\cite{yang2023baichuan2} \\
\midrule
Precision       & \texttt{float16}       & \texttt{float16} & \texttt{float16}               \\
Batch Size      & 16           & 16      &    16          \\ 
Optimizer       & AdamW         & AdamW     & AdamW           \\
Adam $(\beta_1,\beta_2)$ & (0.9,0.95) & (0.9,0.95)  & (0.9, 0.95)  \\
Learning Rate   & 3e-4      & 5e-5     & 3e-4         \\
Sequence Length & 1024          & 1024      & 1024            \\
Warmup Step     & 100             & 100       & 100            \\
Decay style     & \texttt{cosine}       & \texttt{cosine} & \texttt{cosine}                \\
Min. Learning Rate &  0 & 0  & 0\\
Weight Decay    & 0  & 0 & 0 \\
LoRA rank       & 16  & 16 &  16   \\
LoRA $\alpha$   & 16  & 16 &  16   \\
LoRA dropout    & 0.05 & 0.05 & 0.05 \\
LoRA modules &   \makecell[c]{\texttt{gate\_proj}\\ \texttt{up\_proj}\\ \texttt{down\_proj}} &  \makecell[c]{\texttt{gate\_proj}\\ \texttt{up\_proj}\\ \texttt{down\_proj}} & \makecell[c]{\texttt{gate\_proj}\\ \texttt{up\_proj}\\ \texttt{down\_proj}}    \\
\bottomrule
\end{tabular}

    }
    \caption{The hyper-parameter configuration for different base LLMs.}
    \label{tab:hyper}
\end{table*}

\subsection{More Implementation details}

Our experiments are conducted on a cloud Linux server with Ubuntu 16.04 operating system. The codes are written in Python 3.10 using the code from huggingface library\footnote{\scriptsize\url{https://huggingface.co/}}. 
The GPU type is the Nvidia Tesla V100 with 32GB GPU memory.  The detailed hyper-parameter settings for training different base LLMs are shown in Table~\ref{tab:hyper}, which mostly follows \citet{lee2023platypus}. We train each sub-model for $3$ epochs and use the following template for fine-tuning, which is borrowed from \citet{taori2023alpaca}:

{\itshape
Below is an instruction that describes a task. Write a response that appropriately completes the request.

\#\#\# Instruction:
\{instruction\}

\#\#\# Response: \{output\}
}

Note that the language modeling loss is only considered for the output part. 

We use the code from Abel\footnote{\scriptsize\url{https://github.com/GAIR-NLP/abel}}~\citep{ethan2023abel}, Open Instruct\footnote{\scriptsize\url{https://github.com/allenai/open-instruct}}, Language Model Evaluation Harness\footnote{\scriptsize\url{https://github.com/EleutherAI/lm-evaluation-harness}}~\citep{gao2023framework}, Bigcode Evaluation Harness\footnote{\scriptsize\url
{https://github.com/bigcode-project/bigcode-evaluation-harness}}~\citep{allal2022bigcode} and Open Compass~\citep{opencompass2023} for evaluation.

\section{More Experiment Results and Analysis}



\subsection{More Observations on Alignment Tax}
\begin{figure}
    \centering
    \includegraphics[width=1.0\linewidth]{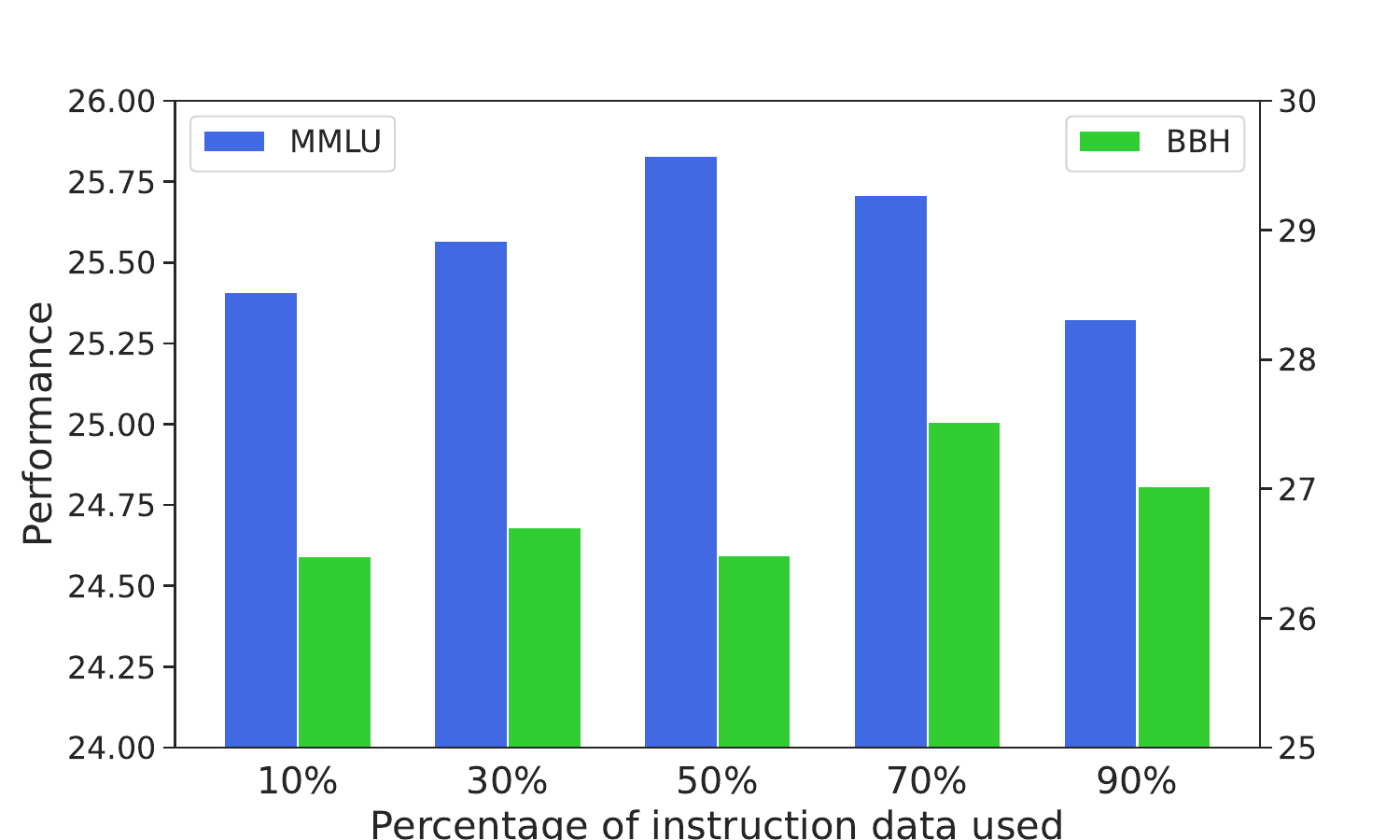}
    \caption{The performance on MMLU (5-shot, accuracy) and BBH (3-shot, exact match) when tuning tinyllama-1.1b with different sizes of instruction-following data from \tulu-V2-mix.}
    \label{fig:tulu-1.1b-updown}
\end{figure}


Our pilot study reveals the decline in MMLU and BBH when tuning Llama-2-7b and Llama-2-13b on the \tulu-V2-mix. To have a more comprehensive understanding, we supplement the experiment on tinyllama-1.1b~\citep{zhang2024tinyllama}, and the experimental results are shown in Table~\ref{fig:tulu-1.1b-updown}.  

\subsection{The effect of the number of sub-models}

\begin{figure}
    \centering
    \includegraphics[width=1.0\linewidth]{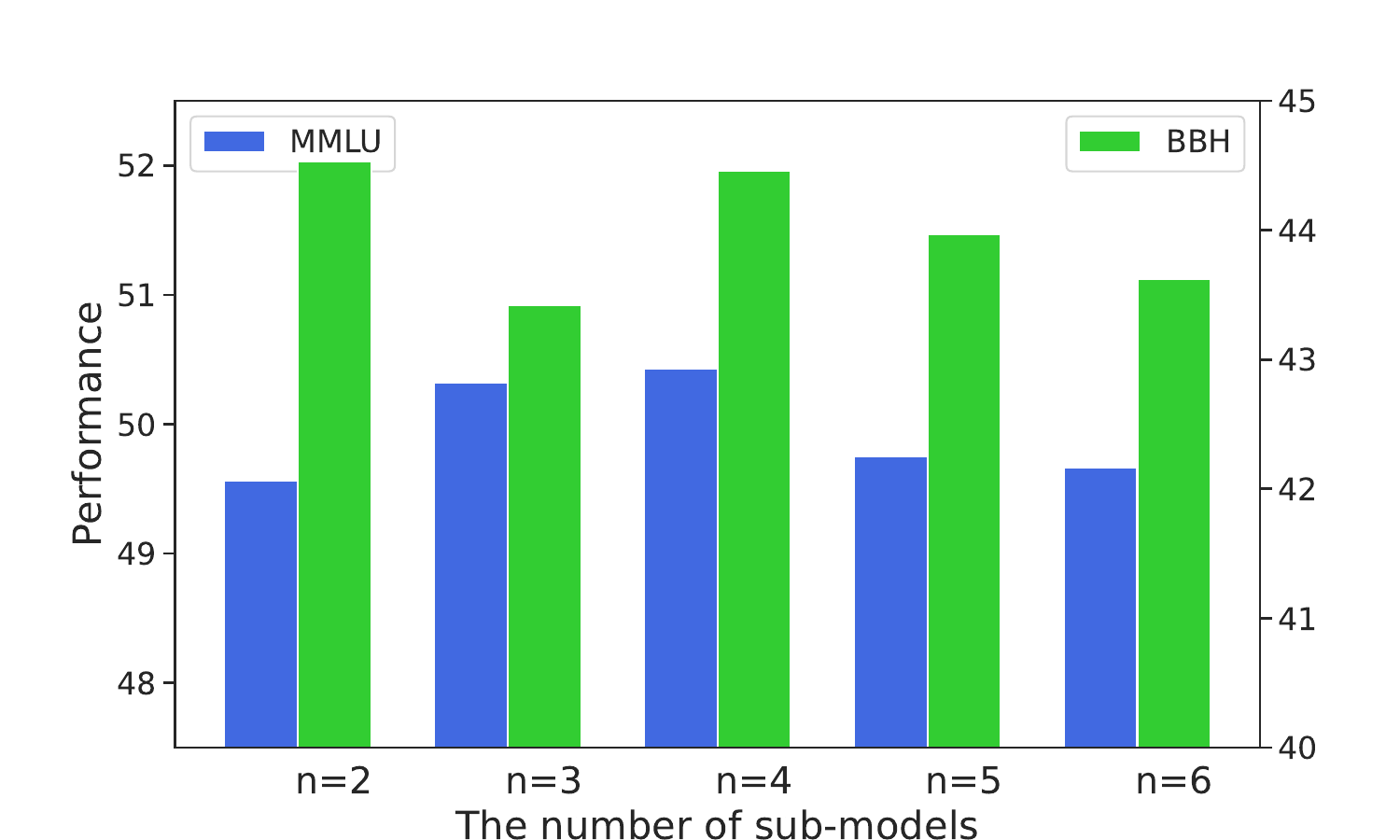}
    \caption{The performance on MMLU (5-shot, accuracy) and BBH (3-shot, exact match) v.s. the number of sub-models}
    \label{fig:n_cluster_2bar}
\end{figure}

To investigate how different choices of $K$ (the number of clusters and the number of sub-models) affect the effectiveness of the \modelname framework, we vary the hyper-parameter $K$ from $2$ to $6$ and the experiment results with different numbers of sub-models are shown in Figure~\ref{fig:n_cluster_2bar}. From the figure we find that $K=4$ attains the best performance among different choices of $K$. We gauge that there exists a trade-off between the acquisition of common knowledge and the forgetting of biases. If $K$ is too small, the data-specific biases are not adequately dispersed. Thus the biases learned by each sub-model are too similar to be forgotten via merging. On the other hand, if $K$ is too large, the average number of samples in each cluster narrows down and probably can not provide sufficient knowledge of instruction-following.  


\label{sec:appendix}

This is an appendix.

\end{document}